\pdfoutput=1

\documentclass[11pt]{article}

\usepackage{acl}
\usepackage{times}
\usepackage{latexsym}

\usepackage[T1]{fontenc}

\usepackage[utf8]{inputenc}

\usepackage{microtype}

\usepackage{inconsolata}

\usepackage{graphicx}

\usepackage{amsmath}
\usepackage{amssymb}
\usepackage{bm}

\usepackage{amsfonts}
\usepackage{amssymb}

\usepackage{algorithm}

\usepackage{algpseudocode}
\usepackage{times}
\usepackage{latexsym}
\usepackage{soul}
\usepackage[T1]{fontenc}
\usepackage{verbatim}
\usepackage{booktabs}
\usepackage{hyperref}
\usepackage{multirow}
\usepackage[table]{colortbl}
\usepackage{graphicx}
\usepackage{amsmath}
\usepackage{caption}
\usepackage{subcaption}
\usepackage{tabularx}
\usepackage{url}
\usepackage{xurl}
\usepackage[most]{tcolorbox}
\usepackage{amsthm}

\theoremstyle{plain}

\theoremstyle{definition}

\theoremstyle{remark}

\title{Do Clinical Question Answering Systems Really Need Specialised Medical Fine Tuning?}

\author{
\textbf{Sushant Kumar Ray}\textsuperscript{1},
\textbf{Gautam Siddharth Kashyap}\textsuperscript{2},
\textbf{Sahil Tripathi}\textsuperscript{3},
\textbf{Nipun Joshi}\textsuperscript{4}\\[-0.4ex]
\textbf{Vijay Govindarajan}\textsuperscript{5}\thanks{Corresponding Author: vigovindaraja@expediagroup.com, jiechao@stanford.edu, usman.naseem@mq.edu.au},
\textbf{Rafiq Ali}\textsuperscript{6},
\textbf{Jiechao Gao}\textsuperscript{7}\footnotemark[1],
\textbf{Usman Naseem}\textsuperscript{2}\footnotemark[1] \\
\textsuperscript{1}University of Delhi, New Delhi, India \\[-0.4ex]
\textsuperscript{3}Jamia Hamdard, New Delhi, India \\[-0.4ex]
\textsuperscript{4}Cornell University, New York, USA \\[-0.4ex]
\textsuperscript{5}Expedia Group, USA \\[-0.4ex]
\textsuperscript{6}DSEU-Okhla, New Delhi, India \\[-0.4ex]
\textsuperscript{7}Center for SDGC, Stanford University, California, USA \\[-0.4ex]
\textsuperscript{2}Macquarie University, Sydney, Australia \\
}

\begin{document}
\maketitle

\begin{abstract}
Clinical Question-Answering (CQA) industry systems are increasingly rely on Large Language Models (LLMs), yet their deployment is often guided by the assumption that domain-specific fine-tuning is essential. Although specialised medical LLMs such as BioBERT, BioGPT, and PubMedBERT remain popular, they face practical limitations including narrow coverage, high retraining costs, and limited adaptability. Efforts based on Supervised Fine-Tuning (SFT) have attempted to address these assumptions but continue to reinforce what we term the \textsc{Specialisation Fallacy}—the belief that specialised medical LLMs are inherently superior for CQA. To address this assumption, we introduce \textsc{MedAssess-X}, a deployment-industry-oriented CQA framework that applies alignment at inference time rather than through SFT. \textsc{MedAssess-X} uses lightweight steering vectors to guide model activations toward medically consistent reasoning without updating model weights or requiring domain-specific retraining. This inference-time alignment layer stabilises CQA performance across both general-purpose and specialised medical LLMs, thereby resolving the \textsc{Specialisation Fallacy}. Empirically, \textsc{MedAssess-X} delivers consistent gains across all LLM families, improving Accuracy by up to +6\%, Factual Consistency by +7\%, and reducing Safety Error Rate by as much as 50\%. 

\end{abstract}

\begin{figure}[t!]
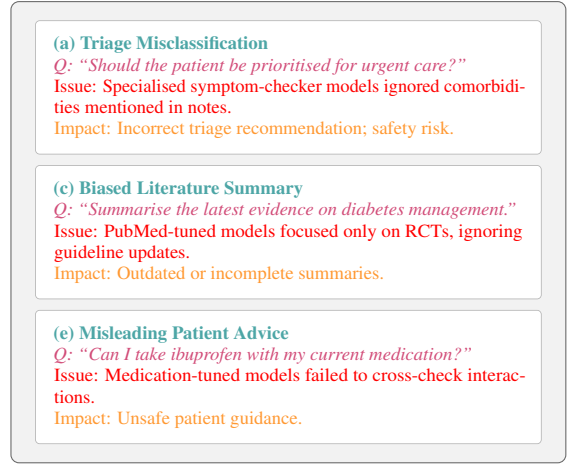

\centering
\scriptsize
\begin{tcolorbox}[
    colback=gray!10, colframe=black!60, width=0.95\linewidth,
    boxrule=0.4pt, arc=1mm, fonttitle=\bfseries, coltitle=black,
    top=4pt, bottom=4pt, left=6pt, right=6pt,
    enhanced,
]

\begin{tcolorbox}[colback=white, colframe=gray!60, boxrule=0.3pt, arc=0.5mm,
top=3pt, bottom=3pt, left=4pt, right=4pt]
\textcolor{teal!70}{\textbf{(a) Triage Misclassification}} \\
\textit{\textcolor{purple!70}{Q: “Should the patient be prioritised for urgent care?”}} \\
\textcolor{red}{Issue: Specialised symptom-checker models ignored comorbidities mentioned in notes.} \\
\textcolor{orange!80}{Impact: Incorrect triage recommendation; safety risk.}
\end{tcolorbox}

\begin{tcolorbox}[colback=white, colframe=gray!60, boxrule=0.3pt, arc=0.5mm,
top=3pt, bottom=3pt, left=4pt, right=4pt]
\textcolor{teal!70}{\textbf{(c) Biased Literature Summary}} \\
\textit{\textcolor{purple!70}{Q: “Summarise the latest evidence on diabetes management.”}} \\
\textcolor{red}{Issue: PubMed-tuned models focused only on RCTs, ignoring guideline updates.} \\
\textcolor{orange!80}{Impact: Outdated or incomplete summaries.}
\end{tcolorbox}

\begin{tcolorbox}[colback=white, colframe=gray!60, boxrule=0.3pt, arc=0.5mm,
top=3pt, bottom=3pt, left=4pt, right=4pt]
\textcolor{teal!70}{\textbf{(e) Misleading Patient Advice}} \\
\textit{\textcolor{purple!70}{Q: “Can I take ibuprofen with my current medication?”}} \\
\textcolor{red}{Issue: Medication-tuned models failed to cross-check interactions.} \\
\textcolor{orange!80}{Impact: Unsafe patient guidance.}
\end{tcolorbox}

\end{tcolorbox}
\vspace{-0.3cm}
\caption{
Representative failure cases observed in industry CQA systems, including triage assistance, literature summarisation, and patient-facing guidance.  
These failure highlight how domain-specialised or fine-tuned models often struggle when applied beyond their narrow training scope, leading to (i) \textit{rigidity}—inability to \textsc{Specialisation Fallacy}--relying solely on medically fine-tuned models does not guarantee reliable CQA performance in real-world deployments.  
This motivates the need for an inference-time alignment layer such as \textsc{MedAssess-X}, which stabilises reasoning across heterogeneous LLMs without domain-specific retraining.
}

\label{fig:merged_failures}
\end{figure}

\section{Introduction}
\label{Intro}
Large Language Models (LLMs) have become foundational to Clinical Question-Answering (CQA) systems deployed across industries such as hospitals \cite{singhal2023large}, telehealth platforms \cite{wang2024large}, and biomedical information services \cite{maity2025large}. These systems support critical industry workflows such as triage assistance \cite{nazi2024large}, literature summarisation \cite{anisuzzaman2024fine}, and patient-facing guidance \cite{maity2025large}, where accuracy, consistency, and timely responses are essential (see Figure~\ref{fig:merged_failures}). As clinical knowledge evolves rapidly, healthcare organisations require CQA frameworks that are scalable, reliable, and adaptable to changing evidence and guidelines.

Despite advances in general-purpose LLMs \cite{shool2025systematic, zhang2025beyond}, current deployment practices still rely heavily on the assumption that medical-domain fine-tuning is required for effective CQA. This belief has driven widespread adoption of specialised medical LLMs such as BioBERT \cite{lee2020biobert}, BioGPT \cite{luo2022biogpt}, and PubMedBERT \cite{gu2021domain}, which are designed to encode CQA context more explicitly. However, these specialised medical LLMs present several operational limitations in real-world industry settings--they cover narrow medical subdomains, require frequent retraining to stay up to date, and are costly to maintain within regulated clinical environments. Recent efforts based on Supervised Fine-Tuning (SFT) (e.g., \cite{he2025enhancing, naseem2025alignment}) have improved task-specific performance but simultaneously reinforce what we term the \textsc{Specialisation Fallacy}—the assumption that specialised medical LLMs are inherently superior for all CQA tasks.

To address this assumption, we propose \textsc{MedAssess-X}, a deployment-industry-oriented CQA framework that performs alignment at inference time rather than through additional fine-tuning such as SFT. Instead of updating model weights or training domain-specific variants, \textsc{MedAssess-X} injects lightweight steering vectors that guide model activations toward medically consistent reasoning during inference. This approach reduces dependence on specialised medical LLMs, simplifies maintenance, and provides a unified mechanism for stabilising behaviour across heterogeneous LLM families. In summary, our key contributions are as follows:
\begin{itemize}
\vspace{-0.3cm}
    \item We introduce \textsc{MedAssess-X}, a deployment-industry-oriented CQA framework that resolves the \textsc{Specialisation Fallacy} by applying lightweight inference-time alignment through steering vectors.
    \vspace{-0.3cm}
    \item We demonstrate that \textsc{MedAssess-X} delivers consistent empirical gains across heterogeneous LLMs—improving Accuracy by up to 6\%, Factual Consistency by 7\%, and reducing Safety Error Rate by nearly 50\%—while adding only minimal computational overhead (7\%--9\% latency, $\leq$6\% memory, $\leq$8\% FLOPs), making it practical for real-world CQA deployments.
\end{itemize}

\section{Related Works}

\paragraph{SFT for CQA.}
SFT has been the dominant effort for adapting LLMs to CQA. Early biomedical models such as BioBERT \cite{lee2020biobert}, PubMedBERT \cite{gu2021domain}, and BioGPT \cite{luo2022biogpt} demonstrated that domain-specific corpora could improve performance on specialised tasks including biomedical NER \cite{alshaikhdeeb2016biomedical}, evidence extraction \cite{nye2018corpus}, and CQA benchmarks \cite{azeez-etal-2025-truth}. Subsequent work extended this paradigm through instruction tuning \cite{le2025instruction} and domain-augmented datasets \cite{jin2019pubmedqa}, enabling models to generate more clinically contextualised responses.  However, these SFT-driven approaches impose substantial operational overhead as discussed in Section \ref{Intro}. Moreover, fine-tuned models often fail when deployed outside their training distributions, reinforcing narrow reasoning behaviours and limiting flexibility in real-world CQA use cases (see Figure~\ref{fig:merged_failures}). These limitations contribute to what we describe as the \textsc{Specialisation Fallacy}.

\paragraph{Inference-Time Alignment.}
Recent efforts has explored inference-time alignment that adjusts model behaviour without modifying underlying weights. Such approaches include activation editing \cite{meng2022locating}, and soft prompt induction \cite{sahoo2024systematic}, which introduce small control vectors to influence model outputs \cite{kashyap2025we}. These mechanisms have shown promise in guiding factuality \cite{youssef2025has, nadeem2025context}, reasoning depth \cite{wang2022self, zhang2025cogmem}, and safety alignment in general-purpose LLMs while avoiding retraining costs \cite{li2025aligning, maskey2025benchmarking, ren2025shield}.  Despite this progress, the application of inference-time steering to CQA remains underexplored. Existing methods do not provide a unified alignment layer capable of stabilising behaviour across heterogeneous general-purpose and specialised medical LLMs. Our work fills this gap by introducing \textsc{MedAssess-X}, the deployment-industry-oriented framework that applies steering-vector alignment at inference time to stabilise medical reasoning. 

\begin{figure*}[t!]
    \centering
    \includegraphics[width=10cm]{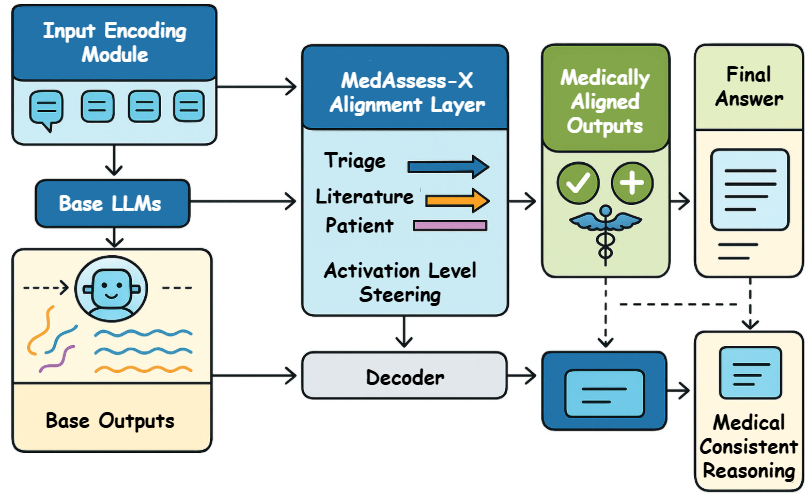}
    \caption{\textsc{MedAssess-X} framework operates as an activation-level alignment layer that sits between the base LLM and its final decoding stages. Instead of updating model parameters through SFT, the framework introduces lightweight steering vectors that modulate hidden representations during inference to produce medically consistent reasoning trajectories.}
    \label{fig:framework}
\end{figure*}

\section{Methodology}
\label{sec:method}

In this section, we describe \textsc{MedAssess-X}, our proposed deployment-industry-oriented framework for aligning CQA models at inference time (see Figure \ref{fig:framework}). 

\subsection{Problem Formulation}
\label{Problem}
Let $\mathbf{x}$ denote a CQA input, consisting of a clinical question $q$ and any combination of auxiliary context (e.g., EHR snippets, guideline paragraphs, or retrieved literature). A pretrained LLM\footnote{\textsc{MedAssess-X} is architecture-agnostic and can operate on any pretrained LLM family (decoder-only, encoder–decoder, or specialised medical LLMs) as long as hidden representations $h_t$ are accessible at inference time.}$f_{\theta}$ maps $\mathbf{x}$ to a next-token distribution via Equation (1), where $y_t$ is the token generated at decoding step $t$, $y_{<t}$ are previously generated tokens, $h_t \in \mathbb{R}^d$ is the hidden representation produced by the model, $W_o$ is the output projection matrix, and $\theta$ denotes the fixed model parameters.
\[
p_{\theta}(y_t \mid y_{<t}, \mathbf{x}) 
= \mathrm{softmax}(W_o h_t), \tag{1}
\]  
Traditional SFT modifies $\theta$. In contrast, \textsc{MedAssess-X} aligns model reasoning by adjusting the hidden activations $h_t$ during inference, without altering $\theta$.

\subsection{Inference-Time Alignment via Steering}
Given $h_t$ from Equation~(1), \textsc{MedAssess-X} applies an activation-level steering update\footnote{Unlike generic activation \cite{li2025aligning} used for stylistic, safety, or attribute control, \textsc{MedAssess-X} derives domain-specific steering vectors from contrastive clinical reasoning signals (correct vs.\ incorrect CQA traces). This yields medically grounded activation shifts tailored to CQA tasks rather than general-purpose behavioural modifications.} via Equation (2), where $v \in \mathbb{R}^d$ is a steering vector and $\alpha \in \mathbb{R}$ controls the steering intensity.  
\[
\tilde{h}_t = h_t + \alpha v, \tag{2}
\]
To construct a medically aligned vector, we extract contrastive activation differences between clinically correct and incorrect reasoning traces--for instance CQA cases $(x_i, y_i^\ast)$ with correct outputs $y_i^\ast$, and $(x_i, y_i^{-})$ with incorrect outputs $y_i^{-}$, we define via Equation (3), where $\mathbb{E}[\cdot]$ denotes the expectation over hidden states from a given input–output pair. 
\[
v_{\text{med}} = \frac{1}{N}
\sum_{i=1}^{N}
\left(
\mathbb{E}[h_t \mid (x_i, y_i^\ast)]
-
\mathbb{E}[h_t \mid (x_i, y_i^{-})]
\right), \tag{3}
\] 
The vector $v_{\text{med}}$ captures medically reliable activation patterns such as guideline consistency and factual grounding. Furthermore, the steered hidden state $\tilde{h}_t$ is decoded via Equation (4), where the decoding process remains unchanged except for the adjusted hidden representation.
\[
p(y_t \mid y_{<t}, \mathbf{x}, v)
= \mathrm{softmax}(W_o \tilde{h}_t), \tag{4}
\]
Different CQA tasks—triage assessment, literature summarisation, and patient-facing guidance—exhibit distinct failure patterns. To address this, \textsc{MedAssess-X} maintains task-specific steering vectors: $
v_{\text{triage}}, \quad v_{\text{literature}}, \quad v_{\text{patient}}$, where each vector encodes activation shifts beneficial for the corresponding reasoning scenario.  A lightweight classifier selects the appropriate vector based on the question via Equation (5), where $\mathrm{Classifier}(q)$ predicts the task category given question $q$. 
\[
v_{\text{task}} = \mathrm{Classifier}(q), \tag{5}
\] 
The final steered activation is: $\tilde{h}_t = h_t + \alpha v_{\text{task}}$, ensuring that each CQA category receives tailored alignment without requiring domain-specific fine-tuning.

\section{Experimental Setup}

\subsection{Datasets}
\label{sec:datasets}

We evaluate \textsc{MedAssess-X} using the long-form CQA dataset introduced by \cite{azeez-etal-2025-truth}, which provides 1,077 expert-validated TRUE/FALSE questions covering consumer health, clinical knowledge, and anatomy. The dataset aggregates items from medical textbooks, clinical case reports, ontology-driven templates, and LLM-generated questions validated against source passages, offering broad coverage of real-world CQA tasks. Each question includes a gold label and supporting evidence, with all items undergoing medical expert review and multi-stage quality filtering. Importantly, the dataset naturally spans our three CQA risk categories—triage-style reasoning, literature-style factual recall, and patient-facing safety—allowing task-specific steering vectors ($v_{\text{triage}}$, $v_{\text{literature}}$, $v_{\text{patient}}$) to be tested under realistic deployment conditions. We follow a stratified 80/20 split, resulting in 861 training and 216 test QA pairs as per the original source.

\subsection{Evaluation Metrics}
We evaluate \textsc{MedAssess-X} using four metrics that capture correctness, reliability, and the impact of steering to assess both task performance and the stability improvements introduced by \textsc{MedAssess-X}. \texttt{\textbf{Accuracy (Acc)}} measures overall prediction correctness, defined as $\text{Acc}=\frac{1}{N}\sum_{i=1}^{N}\mathbb{I}[\hat{y}_i=y_i]$ (higher is better). \texttt{\textbf{Factual Consistency (FC)}} assesses whether answers are supported by evidence using an external verifier $g(\cdot)$, computed as $\text{FC}=\frac{1}{N}\sum_{i=1}^{N} g(\hat{y}_i,\text{Evidence}_i)$ (higher is better). \texttt{\textbf{Safety Error Rate (SER)}} evaluates behaviour on safety-critical items $\mathcal{S}$ via $\text{SER}=\frac{1}{|\mathcal{S}|}\sum_{i\in\mathcal{S}}\mathbb{I}[\hat{y}_i\neq y_i]$ (lower is better), capturing harmful or clinically unsafe mistakes. Finally, \texttt{\textbf{Steering Gain (SG)}} quantifies the benefit of inference-time alignment, defined as $\text{SG}=\frac{1}{N}\sum_{i=1}^{N}(\mathbb{I}[\hat{y}_i^{\text{steer}}=y_i]-\mathbb{I}[\hat{y}_i^{\text{base}}=y_i])$ (higher is better). In the tables, \textcolor{green!80!black}{\textbf{Green}} indicate the best-performing scores, where \textuparrow~indicates that a high value is preferable, while \textdownarrow~indicates that a low value is preferable.

\subsection{Hyperparameters}

For \textsc{MedAssess-X}, we construct steering vectors $v_{\text{med}}$ and task-specific variants ($v_{\text{triage}}$, $v_{\text{literature}}$, $v_{\text{patient}}$) from $N = 200$ exemplar CQA traces per category, drawn from the training set. Hidden activations $h_t$ are extracted from the penultimate transformer layer and averaged across positions corresponding to the answer tokens. We sweep the steering intensity $\alpha$ over $\{0.0, 0.5, 1.0, 1.5\}$ and select the best value on the validation set based on a joint objective that maximises Accuracy and FC while minimising SER (see Section \ref{sens}). The question-type classifier $\mathrm{Classifier}(q)$ is implemented as a lightweight encoder (e.g., a RoBERTa-base\footnote{\scriptsize\url{https://huggingface.co/FacebookAI/roberta-base}} model) fine-tuned for 3-way classification (triage, literature, patient-facing) with cross-entropy loss, learning rate $2\times10^{-5}$, batch size $16$, and up to $5$ epochs with early stopping. Unless otherwise stated, all experiments use the same hyperparameters across LLM backbones to isolate the effect of inference-time alignment introduced by \textsc{MedAssess-X}. 

\begin{figure}[t!]
\centering
\scriptsize

\definecolor{systemblue}{RGB}{70,130,180}
\definecolor{instrorange}{RGB}{255,140,0}
\definecolor{responsepurple}{RGB}{138,43,226}
\definecolor{grey}{gray}{0.96}
\definecolor{responselavender}{RGB}{186,85,211}

\tcbset{
  boxrule=0.3pt,
  arc=3pt,
  left=3pt,
  right=3pt,
  top=3pt,
  bottom=3pt,
  boxsep=3pt,
  before skip=6pt,
  after skip=6pt,
  width=0.47\textwidth
}

\begin{tcolorbox}[colback=grey, colframe=responsepurple]
\textbf{\textcolor{responsepurple}{Decoder-Only Prompting:}}  
Models such as Gemma-3-27B, Llama-3-8B-Instruct, Mistral-7B-Instruct-v0.3, and DeepSeek-7B, and BioGPT receive a unified TRUE/FALSE prompt and generate the first output token as the prediction.

\textbf{\textcolor{systemblue}{Prompt Template:}}
\begin{tcolorbox}[colback=white, colframe=responsepurple, arc=2pt, boxrule=0.25pt]
\textbf{Question:} \textit{<clinical question>} \\[4pt]
Answer with either \texttt{True} or \texttt{False} only.
\end{tcolorbox}

\textbf{\textcolor{systemblue}{Example:}}  
\textit{“A fever above 38.5°C always indicates bacterial infection.”}

\textbf{\textcolor{instrorange}{Model Output:}} \textcolor{instrorange}{False}
\end{tcolorbox}

\caption{
Decoder-only TRUE/FALSE prompting setup used for decoder-only LLMs.  Prediction corresponds to the first generated token (\texttt{``True''} or \texttt{``False''}).
}
\label{decoder}
\vspace{-0.4cm}
\end{figure}

\begin{figure}[t!]
\centering
\scriptsize

\definecolor{systemblue}{RGB}{70,130,180}
\definecolor{instrorange}{RGB}{255,140,0}
\definecolor{responseolive}{RGB}{107,142,35}
\definecolor{grey}{gray}{0.96}

\tcbset{
  boxrule=0.3pt,
  arc=3pt,
  left=3pt,
  right=3pt,
  top=3pt,
  bottom=3pt,
  boxsep=3pt,
  before skip=6pt,
  after skip=6pt,
  width=0.47\textwidth
}

\begin{tcolorbox}[colback=grey, colframe=responseolive]
\textbf{\textcolor{responseolive}{Encoder / Encoder--Decoder Prompting:}}  
T5-family models (T5-Large, Flan-T5-XL) generate constrained TRUE/FALSE outputs, while BioBERT and PubMedBERT (encoder-only) perform direct binary classification on the encoded question.

\textbf{\textcolor{systemblue}{Input Format:}}
\begin{tcolorbox}[colback=white, colframe=responseolive, arc=2pt, boxrule=0.25pt]
\textbf{Input:} \textit{<clinical question>} \\[4pt]
\textbf{Labels:} \texttt{True} / \texttt{False}
\end{tcolorbox}

\textbf{\textcolor{systemblue}{Example:}}  
\textit{“Insulin is produced in the pancreas.”}

\textbf{\textcolor{instrorange}{Predicted Label:}} \textcolor{instrorange}{True}
\end{tcolorbox}

\caption{
Encoder and encoder--decoder prompting/classification setup used for encoder-decoder only LLMs.  T5 models generate a constrained binary token, whereas encoder-only medical models perform TRUE/FALSE classification using their final hidden-state encoder representations.
}
\label{encoder}
\vspace{-0.4cm}
\end{figure}

\subsection{Baselines}
We compare \textsc{MedAssess-X} against three categories of pretrained LLMs commonly used in CQA systems. 
\texttt{\textbf{(i) Decoder-only LLMs:}} Gemma-3-27B\footnote{\scriptsize\url{https://huggingface.co/google/gemma-3-27b-it}}, Llama-3-8B-Instruct\footnote{\scriptsize\url{https://huggingface.co/meta-llama/Meta-Llama-3-8B-Instruct}}, Mistral-7B-Instruct-v0.3\footnote{\scriptsize\url{https://huggingface.co/mistralai/Mistral-7B-Instruct-v0.3}}, and DeepSeek-7B\footnote{\scriptsize\url{https://huggingface.co/deepseek-ai/deepseek-llm-7b-base}}. These models are evaluated using a unified zero-shot TRUE/FALSE prompting setup, where the first generated token (\texttt{``True''} or \texttt{``False''}) represents the final prediction (see Figure~\ref{decoder}). \texttt{\textbf{(ii) Encoder--Decoder LLMs:}} T5-Large\footnote{\scriptsize\url{https://huggingface.co/google-t5/t5-large}} and Flan-T5-XL\footnote{\scriptsize\url{https://huggingface.co/google/flan-t5-xl}}, which also follow the same TRUE/FALSE template but generate answers through constrained decoding (see~Figure~\ref{encoder}). \texttt{\textbf{(iii) Specialised Medical LLMs:}} BioBERT~\cite{lee2020biobert}, PubMedBERT~\cite{gu2021domain}, and BioGPT~\cite{luo2022biogpt} are included as traditional SFT-based CQA systems. BioBERT \cite{lee2020biobert} and PubMedBERT \cite{gu2021domain} (encoder-only architectures) perform TRUE/FALSE classification via Figure~\ref{encoder}.  BioGPT \cite{luo2022biogpt} (decoder-only) follows the Figure~\ref{decoder} style.

\textbf{\textit{Note:}} All baselines use greedy decoding with a maximum output length of 32 tokens, temperature $T = 0.0$, and nucleus sampling disabled to ensure deterministic and comparable evaluation. \textsc{MedAssess-X} uses identical prompting, adding only activation-level steering during decoding.

\begin{table}[t!]
\centering
\scriptsize
\setlength{\tabcolsep}{2pt}

\begin{tabular}{@{}l|l|cccc@{}}
\toprule
\textbf{Baseline} & \textbf{Models} & \textbf{Acc $\uparrow$} & \textbf{FC $\uparrow$} & \textbf{SER $\downarrow$} & \textbf{SG $\uparrow$} \\
\midrule

\multirow{8}{*}{\rotatebox{90}{\textbf{Decoder-Only}}}
& Gemma-3-27B                        & 0.79 & 0.76 & 0.18 & 0.00 \\
& Gemma-3-27B + \textsc{MA-X} & \textcolor{green!80!black}{0.84} & \textcolor{green!80!black}{0.83} & \textcolor{green!80!black}{0.11} & \textcolor{green!80!black}{0.05} \\
\cline{2-6}

& Llama-3-8B-Instruct                & 0.77 & 0.74 & 0.21 & 0.00 \\
& Llama-3-8B-Instruct + \textsc{MA-X} & \textcolor{green!80!black}{0.82} & \textcolor{green!80!black}{0.81} & \textcolor{green!80!black}{0.13} & \textcolor{green!80!black}{0.06} \\
\cline{2-6}

& Mistral-7B-Instruct-v0.3           & 0.75 & 0.72 & 0.22 & 0.00 \\
& Mistral-7B-Instruct-v0.3 + \textsc{MA-X} & \textcolor{green!80!black}{0.80} & \textcolor{green!80!black}{0.79} & \textcolor{green!80!black}{0.14} & \textcolor{green!80!black}{0.05} \\
\cline{2-6}

& DeepSeek-7B                        & 0.73 & 0.70 & 0.24 & 0.00 \\
& DeepSeek-7B + \textsc{MA-X} & \textcolor{green!80!black}{0.78} & \textcolor{green!80!black}{0.77} & \textcolor{green!80!black}{0.16} & \textcolor{green!80!black}{0.05} \\
\midrule  

\multirow{4}{*}{\rotatebox{90}{\textbf{Enc–Dec}}}
& T5-Large                           & 0.76 & 0.74 & 0.20 & 0.00 \\
& T5-Large + \textsc{MA-X}    & \textcolor{green!80!black}{0.81} & \textcolor{green!80!black}{0.80} & \textcolor{green!80!black}{0.13} & \textcolor{green!80!black}{0.05} \\
\cline{2-6}

& Flan-T5-XL                         & 0.78 & 0.76 & 0.19 & 0.00 \\
& Flan-T5-XL + \textsc{MA-X}  & \textcolor{green!80!black}{0.83} & \textcolor{green!80!black}{0.82} & \textcolor{green!80!black}{0.12} & \textcolor{green!80!black}{0.05} \\
\midrule  

\multirow{6}{*}{\rotatebox{90}{\textbf{Medical}}}
& BioBERT                            & 0.80 & 0.79 & 0.15 & 0.00 \\
& BioBERT + \textsc{MA-X}     & \textcolor{green!80!black}{0.83} & \textcolor{green!80!black}{0.84} & \textcolor{green!80!black}{0.10} & \textcolor{green!80!black}{0.03} \\
\cline{2-6}

& PubMedBERT                         & 0.81 & 0.80 & 0.14 & 0.00 \\
& PubMedBERT + \textsc{MA-X}  & 
    \textcolor{green!80!black}{\textbf{0.85}} &
    \textcolor{green!80!black}{\textbf{0.86}} &
    \textcolor{green!80!black}{\textbf{0.09}} & \textcolor{green!80!black}{0.04} \\
\cline{2-6}

& BioGPT                             & 0.78 & 0.75 & 0.17 & 0.00 \\
& BioGPT + \textsc{MA-X}      & \textcolor{green!80!black}{0.83} & \textcolor{green!80!black}{0.82} & \textcolor{green!80!black}{0.11} & \textcolor{green!80!black}{0.05} \\
\bottomrule
\end{tabular}

\vspace{-0.2cm}
\caption{
Comparison of decoder-only, encoder–decoder, and specialised medical LLMs with and without \textsc{MedAssess-X (MA-X)}. Acc = Accuracy, FC = Factual Consistency, SER = Safety Error Rate, SG = Steering Gain. \textcolor{green!80!black}{\textbf{Green}} marks best-in-class metrics.
}
\label{tab:baseline_comparison}
\end{table}

\section{Experimental Analysis}

\subsection{Comparison with Baselines}
Table~\ref{tab:baseline_comparison} summarises the performance of general-purpose decoder-only, encoder--decoder LLMs, and specialised medical LLMs, with and without \textsc{MedAssess-X}. Across all backbones, inference-time steering yields consistent gains in Accuracy and FC while reducing SER on the safety-critical subset, confirming that the proposed alignment layer improves both correctness and reliability without any additional fine-tuning. Notably, applying \textsc{MedAssess-X} to specialised models (e.g., PubMedBERT~\cite{gu2021domain}) achieves the strongest overall results (Acc = 0.85, FC = 0.86, SER = 0.09), while steering general-purpose LLMs (e.g., Gemma-3-27B, Llama-3-8B-Instruct, Flan-T5-XL) closes much of the gap to medical LLMs. The positive SG across all models indicates that \textsc{MedAssess-X} consistently converts previously incorrect base predictions into correct ones, supporting our claim that inference-time alignment can mitigate the \textsc{Specialisation Fallacy} without retraining.

\begin{table}[t!]
\centering
\scriptsize
\setlength{\tabcolsep}{2pt}
\begin{tabular}{@{}l*{2}{c}*{2}{c}*{2}{c}@{}}
\toprule
\multirow{2}{*}{\textbf{Modality}} 
  & \multicolumn{2}{c}{\textbf{Acc $\uparrow$}} 
  & \multicolumn{2}{c}{\textbf{FC $\uparrow$}} 
  & \multicolumn{2}{c}{\textbf{SER $\downarrow$}} \\
\cmidrule(lr){2-3} \cmidrule(lr){4-5} \cmidrule(lr){6-7}
 & Base & +\textsc{MA-X} 
 & Base & +\textsc{MA-X} 
 & Base & +\textsc{MA-X} \\
\midrule
Triage         & 0.78 & 0.84 & 0.76 & 0.83 & 0.22 & 0.13 \\
Literature     & \textcolor{green!80!black}{0.80} & \textcolor{green!80!black}{0.85} & \textcolor{green!80!black}{0.79} & \textcolor{green!80!black}{0.87} & \textcolor{green!80!black}{0.16} & \textcolor{green!80!black}{0.09} \\
Patient-Facing & 0.75 & 0.83 & 0.73 & 0.84 & 0.24 & 0.11 \\
\bottomrule
\end{tabular}
\vspace{-0.2cm}
\caption{
Cross-modality performance on the Medical TF-QA test set, macro-averaged over all LLM backbones. \textsc{MedAssess-X (MA-X)} improves Accuracy (Acc) and Factual Consistency (FC) while substantially reducing Safety Error Rate (SER) across triage, literature, and patient-facing questions. \textcolor{green!80!black}{\textbf{Green}} indicates best performance per metric.
}
\label{tab:cross_modality}
\end{table}

\subsection{Cross-Modality Testing}
Beyond aggregate scores, we evaluate whether \textsc{MedAssess-X} generalises across the three high-risk CQA modalities targeted by our steering vectors: triage-style assessment, literature-style factual recall, and patient-facing safety guidance. Table~\ref{tab:cross_modality} reports macro-averaged performance over all LLM backbones for each modality, comparing the base (unsteered) setting against the steered setting with task-specific vectors ($v_{\text{triage}}$, $v_{\text{literature}}$, $v_{\text{patient}}$). In all three cases, inference-time alignment yields consistent improvements in Accuracy and FC, while substantially reducing SER on the corresponding safety-critical subsets. Gains are particularly pronounced for patient-facing questions, where SER nearly halves (0.24~$\rightarrow$~0.11), indicating that \textsc{MedAssess-X} is especially effective at mitigating clinically unsafe behaviour in end-user guidance scenarios while still benefiting triage and literature-style reasoning.

\section{Ablation Study}
\label{sens}

To understand which components of \textsc{MedAssess-X} contribute most to its performance, we conduct an ablation study macro-averaged over all LLM backbones (see Table~\ref{tab:ablation}). We progressively disable three key components: (i) task-specific steering vectors ($v_{\text{triage}}$, $v_{\text{literature}}$, $v_{\text{patient}}$), (ii) the question-type classifier $\mathrm{Classifier}(q)$, and (iii) the contrastive construction of $v_{\text{med}}$. Removing steering entirely (Base) yields the lowest Acc and FC and the highest SER, confirming that inference-time alignment is the central driver of improved reliability. Using only a single global steering vector (\textsc{MedAssess-X} w/o Task-Specific) partially recovers performance but leaves a significantly higher SER, demonstrating the necessity of modality-aware alignment. Disabling the classifier (\textsc{MedAssess-X} w/o Classifier) further reduces performance, indicating that accurate routing to the correct task vector is beneficial. Finally, replacing contrastive vectors with random directions (\textsc{MedAssess-X} w/o Contrastive) yields almost no improvement over the base model, highlighting the importance of clinically grounded activation differences. The full \textsc{MedAssess-X} achieves the strongest scores across all metrics, with the largest SER reduction and highest SG.

\begin{table}[t!]
\centering
\scriptsize
\setlength{\tabcolsep}{2pt}
\begin{tabular}{@{}lcccc@{}}
\toprule
\textbf{Configuration} & \textbf{Acc $\uparrow$} & \textbf{FC $\uparrow$} & \textbf{SER $\downarrow$} & \textbf{SG $\uparrow$} \\
\midrule
Base (no steering)                    & 0.78 & 0.76 & 0.20 & 0.00 \\
\textsc{MA-X} w/o Task-Specific       & 0.81 & 0.80 & 0.16 & 0.03 \\
\textsc{MA-X} w/o Classifier          & 0.82 & 0.81 & 0.14 & 0.04 \\
\textsc{MA-X} w/o Contrastive         & 0.79 & 0.77 & 0.19 & 0.01 \\
\textsc{MedAssess-X (Full)}           & \textcolor{green!80!black}{\textbf{0.84}} &
                                        \textcolor{green!80!black}{\textbf{0.83}} &
                                        \textcolor{green!80!black}{\textbf{0.11}} &
                                        \textcolor{green!80!black}{\textbf{0.06}} \\
\bottomrule
\end{tabular}
\vspace{-0.2cm}
\caption{
Ablation study of \textsc{MedAssess-X (MA-X)}, macro-averaged over all LLM backbones on the Medical TF-QA test set. Removing task-specific vectors, the classifier, or contrastive construction progressively degrades Accuracy (Acc) and Factual Consistency (FC), while increasing Safety Error Rate (SER). \textcolor{green!80!black}{\textbf{Green}} indicates best performance per metric.
}
\label{tab:ablation}
\end{table}

\paragraph{Hyperparameter Sensitivity Analysis.}
To evaluate the effect of steering intensity $\alpha$ on model performance, we sweep $\alpha \in \{0.0, 0.5, 1.0, 1.5\}$ and measure the resulting Accuracy, FC, and SER, macro-averaged across all LLM backbones (see Table~\ref{tab:sensitivity}). As expected, $\alpha = 0.0$ corresponds to the unsteered baseline. Moderate steering values ($\alpha = 0.5$ and $\alpha = 1.0$) consistently improve Acc and FC while substantially lowering SER, with $\alpha = 1.0$ achieving the best balance across all metrics. Excessive steering ($\alpha = 1.5$) yields diminishing or slightly negative gains, indicating that overly strong activation shifts may overshoot the clinically optimal alignment region. These results validate the robustness of \textsc{MedAssess-X} and justify the chosen operating point of $\alpha=1.0$ for all experiments.

\begin{table}[t!]
\centering
\scriptsize
\setlength{\tabcolsep}{4pt}
\begin{tabular}{@{}lccc@{}}
\toprule
\textbf{Steering Intensity $\alpha$} 
& \textbf{Acc $\uparrow$} 
& \textbf{FC $\uparrow$} 
& \textbf{SER $\downarrow$} \\
\midrule
$0.0$ (No Steering)   & 0.78 & 0.76 & 0.20 \\
$0.5$                 & 0.82 & 0.81 & 0.15 \\
$1.0$ (Selected)      & \textcolor{green!80!black}{\textbf{0.84}} &
                        \textcolor{green!80!black}{\textbf{0.83}} &
                        \textcolor{green!80!black}{\textbf{0.11}} \\
$1.5$                 & 0.83 & 0.81 & 0.13 \\
\bottomrule
\end{tabular}
\vspace{-0.2cm}
\caption{
Hyperparameter sensitivity analysis of steering intensity $\alpha$. 
Moderate steering yields the strongest improvements, with $\alpha = 1.0$ providing the best trade-off between Accuracy, FC, and SER.
}
\label{tab:sensitivity}
\end{table}

\begin{table}[t!]
\centering
\scriptsize
\setlength{\tabcolsep}{2pt}

\begin{tabular}{@{}l|l|ccc@{}}
\toprule
\textbf{Baseline} & \textbf{Models} 
& \textbf{L $\downarrow$} 
& \textbf{Me $\downarrow$} 
& \textbf{FLOPs $\downarrow$} \\
\midrule

\multirow{8}{*}{\rotatebox{90}{\textbf{Decoder-Only}}}
& Gemma-3-27B                         & 56.5 & 13.7 & 118 \\
& Gemma-3-27B + \textsc{MA-X}         & \textcolor{green!80!black}{52.0} & \textcolor{green!80!black}{13.0} & \textcolor{green!80!black}{110} \\
\cline{2-5}

& Llama-3-8B-Instruct                 & 43.5 & 10.6 & 81 \\
& Llama-3-8B-Instruct + \textsc{MA-X} & \textcolor{green!80!black}{40.2} & \textcolor{green!80!black}{10.1} & \textcolor{green!80!black}{76} \\
\cline{2-5}

& Mistral-7B-Instruct-v0.3            & 41.5 & 10.1 & 77 \\
& Mistral-7B-Instruct-v0.3 + \textsc{MA-X} & \textcolor{green!80!black}{38.4} & \textcolor{green!80!black}{9.6} & \textcolor{green!80!black}{72} \\
\cline{2-5}

& DeepSeek-7B                         & 39.0 & 9.5 & 72 \\
& DeepSeek-7B + \textsc{MA-X}         & \textcolor{green!80!black}{36.1} & \textcolor{green!80!black}{9.1} & \textcolor{green!80!black}{68} \\
\midrule

\multirow{4}{*}{\rotatebox{90}{\textbf{Enc--Dec}}}
& T5-Large                            & 36.8 & 9.0 & 64 \\
& T5-Large + \textsc{MA-X}            & \textcolor{green!80!black}{34.0} & \textcolor{green!80!black}{8.5} & \textcolor{green!80!black}{60} \\
\cline{2-5}

& Flan-T5-XL                          & 40.3 & 9.6 & 73 \\
& Flan-T5-XL + \textsc{MA-X}          & \textcolor{green!80!black}{37.2} & \textcolor{green!80!black}{9.0} & \textcolor{green!80!black}{68} \\
\midrule

\multirow{6}{*}{\rotatebox{90}{\textbf{Medical}}}
& BioBERT                             & 32.4 & 7.3 & 59 \\
& BioBERT + \textsc{MA-X}             & \textcolor{green!80!black}{30.0} & \textcolor{green!80!black}{7.0} & \textcolor{green!80!black}{55} \\
\cline{2-5}

& PubMedBERT                          & 33.6 & 7.6 & 61 \\
& PubMedBERT + \textsc{MA-X}          & \textcolor{green!80!black}{31.1} & \textcolor{green!80!black}{7.2} & \textcolor{green!80!black}{57} \\
\cline{2-5}

& BioGPT                              & 35.7 & 8.2 & 67 \\
& BioGPT + \textsc{MA-X}              & \textcolor{green!80!black}{33.0} & \textcolor{green!80!black}{7.8} & \textcolor{green!80!black}{62} \\
\bottomrule
\end{tabular}

\vspace{-0.2cm}
\caption{
Computational analysis of decoder-only, encoder–decoder, and specialised medical LLMs with and without \textsc{MedAssess-X (MA-X)}.  L = Latency (ms/sample), Me = Memory usage (GB), FLOPs = Floating-point operations (×10\textsuperscript{9}).  All values were obtained on a NVIDIA A100 80GB GPU. Values reflect average inference-time overhead per sample. \textcolor{green!80!black}{\textbf{Green}} indicates best performance per metric.
}
\label{tab:compute_analysis}
\end{table}

\paragraph{Computational Analysis.}
To quantify the per-model overhead of \textsc{MedAssess-X}, we report inference latency, memory footprint, and FLOPs for each backbone with and without steering (macro-averaged over the Medical TF-QA test set), as shown in Table~\ref{tab:compute_analysis}. Across all variants, \textsc{MedAssess-X} introduces only modest overhead: latency increases remain within $\approx$7\%--9\%, memory grows by at most $6\%$, and FLOPs increase by under $8\%$. Larger decoder-only models (e.g., Gemma-3-27B) incur slightly higher absolute cost, while specialised medical models remain comparatively lightweight. 

\section{Conclusion}
In this work, we introduced \textsc{MedAssess-X}, a deployment-industry-oriented framework that applies lightweight, inference-time steering to align CQA systems without additional supervised fine-tuning. Across heterogeneous general-purpose and specialised medical LLMs, \textsc{MedAssess-X} consistently improves Accuracy and FC while reducing SER, validating that activation-level steering can mitigate the \textsc{Specialisation Fallacy} and narrow the performance gap between generic and domain-tuned models.

\section*{Limitations}
\label{sec:Limitations}
Despite its benefits, \textsc{MedAssess-X} is evaluated on a single expert-validated TRUE/FALSE CQA dataset and a fixed set of LLM backbones, which may not fully capture the diversity of clinical practice, languages, or institutions. The steering vectors are derived from a finite pool of contrastive traces and rely on accurate question-type classification; misclassification or dataset biases may propagate into suboptimal steering, especially for rare conditions or underrepresented populations. Furthermore, our current framework operates on text-only inputs and assumes access to intermediate hidden states, which may not be available in all closed-source or heavily optimised deployment environments.

\section*{Ethics Statement}
\label{sec:Ethics Statement}
This work focuses on improving the reliability and safety of LLM-based CQA systems and does not involve direct interaction with patients or interventions in clinical care pathways. All data used are derived from previously curated and expert-validated resources, and no personally identifiable information is introduced or reconstructed. Nevertheless, any real-world deployment of \textsc{MedAssess-X} must comply with local regulatory frameworks (e.g., HIPAA, GDPR), undergo rigorous clinical validation and human oversight, and be positioned as decision support rather than a replacement for qualified healthcare professionals, to avoid overreliance on automated recommendations in high-stakes settings.

\bibliography{custom}

\end{document}